\documentclass[review]{elsarticle}

\usepackage{lineno,hyperref}
\usepackage{multirow}
\usepackage{tabularx}
\usepackage{caption}
\usepackage{subcaption}
\usepackage{diagbox}
\usepackage{pgfplots}
\usepackage{longtable}
\usepackage{pgfplotstable}
\usepackage{pdflscape}

\usepackage{cleveref}
\usepackage{makecell}
\usepackage{tikz}
\usetikzlibrary{fit,positioning}

\begin{document}

\begin{frontmatter}

\title{A Comprehensive Review of Trends, Applications and Challenges In Out-of-Distribution Detection}

\author[ferdowsi]{Navid Ghassemi}
\author[ferdowsi]{Ehsan Fazl-Ersi\corref{mycorrespondingauthor}}
\cortext[mycorrespondingauthor]{Corresponding author}
\ead{fazlersi@um.ac.ir}

\address[ferdowsi]{Computer Engineering department, Ferdowsi University of Mashhad, Mashhad, Iran.}

\begin{abstract}
With recent advancements in artificial intelligence, its applications can be seen in every aspect of humans' daily life. From voice assistants to mobile healthcare and autonomous driving, we rely on the performance of AI methods for many critical tasks; therefore, it is essential to assert the performance of models in proper means to prevent damage. One of the shortfalls of AI models in general, and deep machine learning in particular, is a drop in performance when faced with shifts in the distribution of data. Nonetheless, these shifts are always expected in real-world applications; thus, a field of study has emerged, focusing on detecting out-of-distribution data subsets and enabling a more comprehensive generalization. Furthermore, as many deep learning based models have achieved near-perfect results on benchmark datasets, the need to evaluate these models' reliability and trustworthiness for pushing towards real-world applications is felt more strongly than ever. This has given rise to a growing number of studies in the field of out-of-distribution detection and domain generalization, which begs the need for surveys that compare these studies from various perspectives and highlight their straightens and weaknesses. This paper presents a survey that, in addition to reviewing more than 70 papers in this field, presents challenges and directions for future works and offers a unifying look into various types of data shifts and solutions for better generalization.
\end{abstract}


\end{frontmatter}

\section{Introduction}

Arguably, the ultimate goal of any data scientist is to discover the underlying distribution of data through empirical analysis of available samples \cite{bishop2006pattern}. This originates a flaw, the assumption that samples are enough to make decisions on the shape of the underlying distribution. In reality, though, samples can be at best enough to make a model work under a closed-world assumption (an environment where the absence of information is interpreted as negative information \cite{keet2013closed}).

The whole foundation of machine learning (ML) is based on the assumption that identically and independently distributed (i.i.d) data will be available in test time, an assumption that, sadly, can not be held true in many real-world scenarios \cite{dundar2007learning}. Nowadays, ML and deep learning (DL) algorithms are used in every aspect of life. The entire closed-world and i.i.d assumptions are violated when using these algorithms in real-world scenarios. Many algorithms that have already solved their task and even have outperformed human benchmarks fail dramatically on deployment \cite{zhang2021dive}. The reasoning behind their poor performance can be due to changes in data, i.e., distribution shifts. 

The effect of these changes can be seen in all types of applications, such as medical diagnosis \cite{cohen2020limits} or autonomous driving \cite{filos2020can}. Sometimes, the slightest changes in data, such as changes in weather conditions for autonomous cars, can cause the model to perform unacceptable \cite{erkent2020semantic}. The problem can be even more severe when models are faced with new classes of data, and then they make a certain decision about them \cite{geng2020recent}. Even some exploit similar flaws to add small perturbations, not even noticeable by the human eye, to images and mislead models to make wrong decisions \cite{goodfellow2014explaining}.

Research on solving these issues has been going on for some time. Many keywords and work directions have investigated different aspects of these problems and solutions. Interpretable and explainable decisions, Trustworthy AI, Open-set Recognition, Uncertainty analysis, Model Calibration, Domain Generalization, and many other directions have been investigated, aiming to make an AI system capable of withstanding challenges faced in open-world scenarios. Many researchers have already investigated these paths and come up with solutions to help toward this goal.

This field of research still lags behind other directions and aspects of ML. There are well-known benchmarks for making a model for image classification, and nearly all papers and works agree on metrics, notation, and evaluations. However, in distribution shifts, still many papers differ from others in these terms, making their comparison more difficult; thus, more works are considered parallel instead of building on top of another. Lack of a unified definition, notation, and benchmarks are all parts of the problems.

The main objective of this survey is to bring together a review of important works to unify different methodologies and looks at the problem at hand and serve as a starting point for researchers and ultimately and hopefully, pave the path for the creation of robust and trustworthy AI systems. To carry out these goals, we have reviewed more than 70 papers and focused on top-tier venues such as NeruIPS and ICML and prestigious journals to review important works and ideas. The primary objectives of this work are as follows :

\begin{itemize}
    \item Introducing a notation and a graphical model to look at distribution shifts
    \item Reviewing datasets, benchmarks and metrics to create a more plug-and-play mean for researchers in this field
    \item Analysing trends and importance of each research direction in this field
    \item Investigating challenges and possible future directions
\end{itemize}

The rest of the paper is structured as follows. First, the search strategy for finding papers is presented in the next section of the paper. The third section of the paper is devoted to presenting the history and related works; the description of the main ideas of methods is presented here. Next, a section is added on notation and definitions. A review of papers is presented in section five, with section six discussing challenges and future directions based on these papers. Lastly, section seven concludes the paper.

\section{Search Strategy and Review Structure}

To properly form this survey, we categorized our scope of work into four different problems and then searched for papers in each. For every category, a subsection is assigned to investigate different aspects of the problem and its applications. Description of datasets and benchmarks, remarkable ideas, and review of papers are presented here. These subsections are formed as disjoint as possible, so readers can focus on the problem they intended with the least amount of reading possible. Keywords were selected for each problem empirically by analyzing keywords of a few prior works, all of which are listed in subsections of section 5. For each problem, 15-20 papers are reviewed, making a total of 70 papers. Papers are selected in a way to represent essential aspects of their group. To this end, we have selected papers in a way that:

\begin{itemize}
\item Most cited papers (cite/year)
\item Recent papers (2020-2022) 
\item Important and remarkable works
\item Different applications and data type
\end{itemize}

To select a diverse group of papers to represent various ideas, trends, and applications of each category, we have selected them from top-tier venues such as ICML and NeurIPS and prestigious journals. The last search for the paper was conducted on Aug 31st, 2022, and search keywords were selected empirically by analyzing keywords of a few related works for each problem.

\section{Related Work and History}

Analyzing shifts in the distribution of data and finding instances different from the rest of the dataset is not a recent trend, and its footstep can be found in older literature; in different names such as concept drift \cite{widmer1996learning} and concept shift \cite{vorburger2006entropy}, outlier, and novelty detection \cite{scholkopf1999support}, and covariate shift \cite{shimodaira2000improving} to name a few. An influential paper by Moreno-Torres et al. \cite{moreno2012unifying} in 2012 tried to unify the notation used to describe dataset shifts in classification problems and gave a comprehensive analysis of the causes of each type of shift. 

The number of papers and works done to tackle challenges in the field implies how important these issues are, yet the path to solving all these issues is still long. The course of research also dramatically changed in the era of deep learning, as the models could process massive datasets. Moreover, many deep learning methods have already passed human benchmarks, and they are one step closer to being used for real-world applications; however, the nature of distribution shifts in open-world scenarios can hold them back. This has drawn the attention of many researchers in recent years toward this field. 

OOD samples and distribution shifts can occur due to many factors; the first one being changes in the environment, such as changes in weather, e.g., the model is trained on images taken on sunny days, then tested in rainy conditions. Another simple and well-known example is day-vs-night images. Detecting changes also might become harder, such as differences between images produced by imaging devices of different hospitals. As long as these changes are predictable, data augmentation (DA) can help the model to generalize well to them, yet they are not always known, or creating a DA method for them is complicated. One example is the transformation of real images to sketches, which is easily achievable by extracting edges; however, the reverse is significantly harder. In the past few years, some models have achieved extraordinary performances for the unpaired image-to-image classification, such as CycleGAN \cite{zhu2017unpaired} which has shed light on creating working DA methods, but unaccounted changes in the environment still remain an issue in need of handling.

The following reason behind these shifts can be seen in changes in the nature of events. A model trained to classify between different types of brain tumors does not anticipate any data points not to have tumors or to have other types of disease, and they are also generally not evaluated on outlier data points. Nonetheless, when faced with new types of data, they may produce results with high certainty. There are simple methods taken for tackling this challenge, such as introducing outliers as another class to the model, yet the number of possibilities for other events might become countless, particularly for problems such as face recognition.

After events and environmental changes, the most recognized and infamous cause behind OOD samples is noise. Having a source of noise affecting data is generally inevitable, and it is also usually accepted and accounted for in ML literature. Works on detecting outliers have been a field of research for a long, and its methods are usually considered well-developed.

The last cause of change is rooted within the behavior of the agent itself. This is mainly the case where an adversary agent attacks a machine learning algorithm. In adversarial attacks, the agent usually changes the image slightly, which is not noticeable to a human observer, but it changes the prediction of the model entirely. Arguably, these small changes can also be viewed as noises added to images, yet noise comes from a random process, while on the contrary, these perturbations are created with intention and in a specific way to mislead the model; thus we have also given them a category totally for themselves.

Given all these reasons behind changes and shifts in the distribution of data, papers can be categorized into a few groups based on the problem they are tackling. The next section of the paper gives an overview of each problem and reviews works done for that. There are already a few surveys reviewing papers on out-of-distribution detection and distribution shifts \cite{yang2021generalized,shen2021towards,geng2020recent,zhou2021domain,wang2022generalizing,blazquez2021review,xu2020adversarial}, yet these papers mainly focus on one problem deeply, or their literature does not contain more recent trends and publications. Nevertheless, the interested reader is encouraged and also referred to read those surveys, specifically when focusing on one of the problems. 

There are a few approaches one can take when working on the detection and generalization of models for OOD data, a list of them is as follows:

\subsection{Information theory backed models}

Information theory and probability theory provide many tools that are naturally made to test distribution matching. In the simplest form, getting KL divergence between the output of a classifier and uniform distribution or using extreme value theory (EVT) \cite{smith1990extreme} are used to detect OOD samples. In more advanced methods, however, tools such as mutual information (MI) \cite{wang2021learning}, or complexity of data \cite{serra2019input} are seen to be used numerously. This category is also quite popular among researchers, given that their methods can usually be used without further training a neural net.

\subsection{Reconstruction Error}

Taking data into a latent space and then taking it back to the main space is an idea used by autoencoders to learn a robust representation of data in a smaller latent space \cite{goodfellow2016deep}. Reconstruction-based methods work to utilize the idea that models trained on some data will generate huge reconstruction errors on OOD samples. This simple idea has been used on many ML and statistical algorithms, from PCA \cite{jackson2004robust} to variational autoencoders \cite{lin2020anomaly}.

\subsection{Density and Distance-based}

Similar to reconstruction-based methods, these methods also usually benefit from how data are in a latent (or feature) space. Here, the method usually requires some training data to learn how data are distributed; then, OOD samples are detected by finding different behavior from training data in latent space. Some of these methods explicitly work by finding the distance, while others first learn the density of in-samples and then find the divergence from the dense segment of the distribution, yet arguably they both are working similarly. 

\subsection{Generative}
Generative methods mostly became popular in the past few years, courtesy of the outstanding performance of GANs \cite{goodfellow2014generative} and other generative models in image generation. These methods also are intrinsically density-based, given that most generative models aim to learn the underlying distribution of data, whether explicitly or implicitly. Using a discriminator of GAN or using the generator to generate similar in-samples to train the model further are amongst the most popular approaches when it comes to using GANs to detect OOD, but using other generative models such as VAEs is also investigated in research papers.

\subsection{Novel methods}

Researchers in OOD detection literature use various directions and approaches. While the four mentioned category is more commonly seen in papers, many novel ideas do not belong to them. Disentangled representation learning \cite{trauble2021disentangled}, causal learning \cite{wang2022causal}, and concept learning \cite{kardan2021towards} are some examples. Compared to other paradigms, these usually have a more recent model as the backbone of their structure, yet they are limited by the weaknesses of their underlying model, which can be numerous given their newness.

\section{Notation and Definitions}
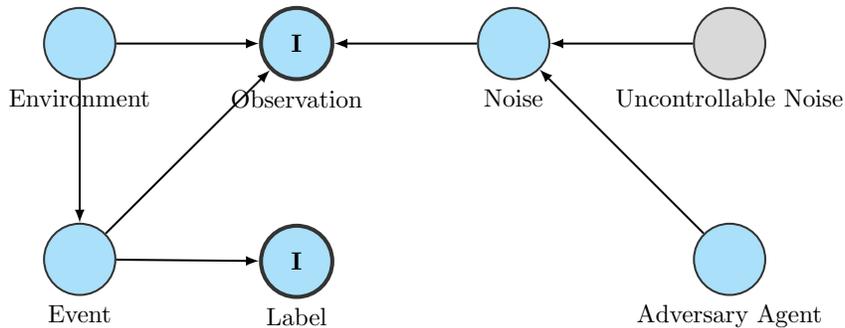
\begin{figure*}
\centering
\resizebox{4.5in}{!}{
\begin{tikzpicture}
\tikzstyle{main}=[circle, minimum size = 10mm, thick, draw =black!80, node distance = 20mm]
\tikzstyle{obsed}=[circle, minimum size = 10mm, ultra thick, draw =black!80, node distance = 20mm]
\tikzstyle{connect}=[-latex, thick]
\tikzstyle{box}=[rectangle, draw=black!100]
  \node[main , fill = cyan!30] (env) [label=below:Environment] { };
  \node[obsed, fill = cyan!30] (ob) [right=of env,label=below:Observation,label=center:\textbf{I}] { };
  \node[main, fill = cyan!30] (no) [right=of ob,label=below:Noise] {};
  \node[main, fill = gray!30] (unno) [right=of no,label=below:Uncontrollable Noise] { };
  \node[main, fill = cyan!30] (event) [below=of env,label=below:Event] { };
  \node[obsed, fill = cyan!30] (label) [below=of ob,label=below:Label,label=center:\textbf{I}] { };
  \node[main, fill = cyan!30] (adv) [below=of unno,label=below:Adversary Agent] { };
  \path (env) edge [connect] (ob)
        (no) edge [connect] (ob)
		(unno) edge [connect] (no)
		(env) edge [connect] (event)
		(event) edge [connect] (ob)
		(event) edge [connect] (label)
		(adv) edge [connect] (no);

\end{tikzpicture}}
\caption{SCM Model For Data\label{SCM}}
\end{figure*}
In this section of the paper, a mathematical look is given to distribution shifts in a search for boundaries between different problem settings. Firstly, we look into how previous works have categorized change, and then we propose a new look to resolve a few issues of prior ones with the help of a structural causal model (SCM) \cite{glymour2016causal}.

\subsection{Distribution Shifts - an Overview}
In this section of the paper, a mathematical overview is presented on the categorization of shifts by prior works. Firstly, as an influential pioneer in defining distribution shifts, we consider \cite{moreno2012unifying} look into how shifts differ from each other, then we attend the issues with this model in a quest for a new one. 

In \cite{moreno2012unifying}, distribution shifts are defined as times where the distribution of data ($P(x,y)$) for train and test are not the same, or in other words, $P_{train}(x,y) \neq P_{test}(x,y)$. As this paper investigated distribution shifts in classification problems, $x$ is defined as data points sampled from an underlying distribution ($x \sim X$), and y is a class label ($y \in Y = \{y_1, y_2, \cdots , y_n\}$); however, this notation can be used for other types of problems such as regression as well, merely by a few changes. To not make things complicated, we merely assume an underlying classification problem and continue with defined notation.

In the paper, problems are categorized into two groups, places where an $X \rightarrow Y$ causal link holds, i.e., credit card fraud detection. The other group is $X \leftarrow Y$, such as medical diagnosis, where the disease causes the symptoms. Given this, the joint distribution can be calculated by $P(X)P(Y|X)$ or $P(Y)P(X|Y)$, respectively. Now, the types of shifts are analyzed:
\begin{itemize}
    \item Covariate shift: Only happens in $X \rightarrow Y$ setting, when $P_{train}(y|x) = P_{test}(y|x)$ but $P_{train}(x) \neq P_{test}(x)$
    \item Prior probability shift: Only happens in $X \leftarrow Y$ setting, when $P_{train}(x|y) = P_{test}(x|y)$ but $P_{train}(y) \neq P_{test}(y)$
    \item Concept shift (concept drift): Happens in both settings, in first by $P_{train}(y|x) \neq P_{test}(y|x)$ when $P_{train}(x) = P_{test}(x)$ and in second when $P_{train}(x|y) \neq P_{test}(x|y)$ when $P_{train}(y) = P_{test}(y)$
    \item Otherwise: where no equality holds true.
\end{itemize}

There are a few issues with this look at data distribution:
\begin{enumerate}
    \item Given the type of problem, one needs to find which setting applies.
    \item Semantic shifts (such as the introduction of a new class) can be well-defined only within the second setting.
    \item When switching to more complected spaces such as images and collecting datasets empirically instead of creating them, holding equality in distribution becomes impossible, and most of the problems come from the last category.
\end{enumerate} 
Given that these shifts are direct consequences of some unobserved variables, we try to make a new look by adding those variables to the causal model.

\subsection{Distribution Shifts - a new look}

Considering the previous view of the causal model behind data, the only variables are $X$ and $Y$, or in other words, observation and label, and the two possible causal links are whether $X \rightarrow Y$ or $X \leftarrow Y$. Here, we try to create a more complicated causal model for data with two goals in mind; first, the direction of causal links should be the same for any types of data, and second, to factor in different causes behind distribution shifts. Figure \ref{SCM} shows our suggested causal model.

Here, we first removed any link between observation and label. By doing this, our structure becomes similar for any type of dataset. Next, we have added a node called the event. This event can be a class label; in other words, the event is what the model aims to recognize from the observation. Next, we need to add another node for the environment, showing the domain or environment effect directly. It is self-evident that the environment causally affects the observation, but also it should be noted that change in environment can cause a change in the distribution of events, even introducing new ones. Next, we have added a node for noise, but the source of noise can be uncontrollable and natural, such as noises in measurement systems, or it can be due to actions of an adversary agent, which is also added to the SCM. 

There are a few points that need to be clarified about this graph as well; first, noise can also be causally related to environment and event, but we have obviated the overhead of considering all those links by making the uncontrollable noise an un-observed node. Next, noise can also affect labels, but that is not within the scope of OOD literature; therefore, we have also removed that relation as well.

Finally, the whole distribution can be viewed as:
\begin{equation}
    P(X,Y,\textcolor{gray}{Env},\textcolor{gray}{Event},\textcolor{gray}{Noise})
\end{equation}
and thus, distribution shifts can occur by a shift in $P(Env)$, $P(Event)$, $P(Noise)$, or a combination of them.

\section{Reviews}
This section of the paper is devoted to reviewing papers, divided based on the problem they try to solve. First, in each subsection, a description of the problem, benchmarks, and evaluation metrics is presented. Then, for each subsection, a table summarizes papers in that problem setting.


\subsection{Domain Generalization}

\begin{figure*}[h]	
\centering
\includegraphics[width=\textwidth]{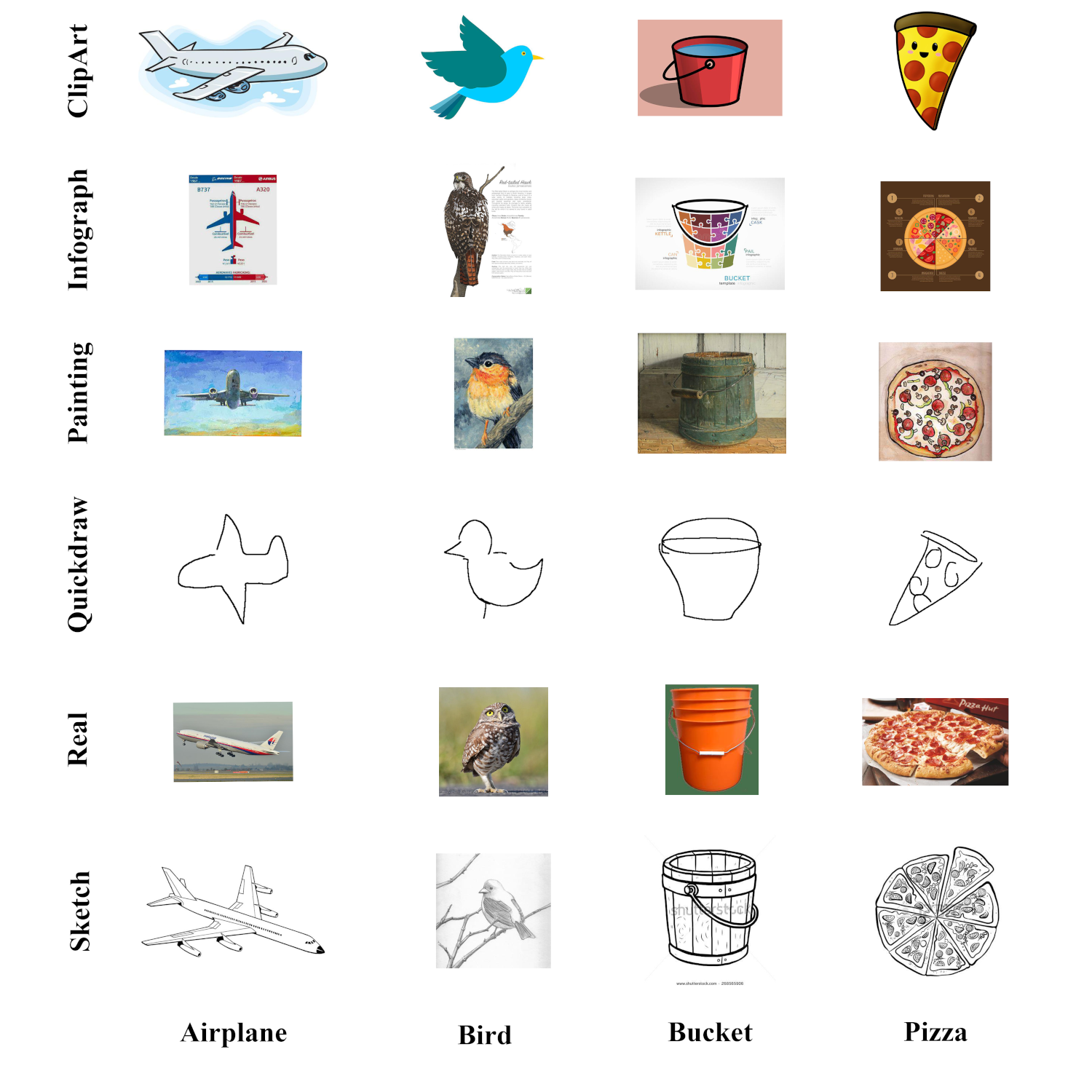}
\caption{Illustration of a few examples from DomainNet dataset.\label{domaingenex}}
\end{figure*}

In this problem setting, mainly the domain (environment) from which the data is sampled changes between training and testing. In other words, changes happen by a shift in $P(Env)$. Figure \ref{domaingenex} shows an example of these types of changes when looking from top to bottom. While the class labels are the same amongst all different domains, some concept shift exists between them.

Image is the most investigated type of data for domain generalization; a few also investigated other types of data such as time-series \cite{hu2021causal}. The reasoning behind this lies within the definition of domain and domain changes; anyone has an intuition of differences when they face a painting in comparison to natural scenery, but the effects of changes in data gathering equipment in different hospitals \cite{roy2022does} is not similar to this. Naturally, datasets are also mainly on the image. Table \ref{domgendata} provides details on some of the main benchmark datasets of this field.

\begin{table}[h] 
\caption{Domain Generalization Datasets.\label{domgendata}}
\centering
\resizebox{3.5in}{!}{
\begin{tabular}{|c|c|c|c|}
\hline
\textbf{Dataset}	& \makecell{\textbf{Number of}\\\textbf{Domains}}	& \makecell{\textbf{Total Number}\\\textbf{of Images}} & \textbf{Description}\\
\hline
RotatedMNIST \cite{ghifary2015domain} & 6 & 6000 & \makecell{Each domain is created\\by rotating MNIST images with\\an specific angle} \\
\hline
ColoredMNIST \cite{arjovsky2019invariant} & 3 & \makecell{60000 (all\\MNIST)} & \makecell{Digits are colored in Green\\or Red with high correlation to\\digits. Ratio of correlation\\is different between environments}\\
\hline
VLCS \cite{fang2013unbiased} & 4 & 10729 & \makecell{Created by selecting images\\from four different dataset.\\Selected classes are : bird,\\car, chair, dog, person.}\\
\hline
PACS \cite{li2017deeper} & 4 & 9991 & \makecell{Domains are: Photo,\\Art Painting, Cartoon,\\ Sketch - selected from\\seven classes}\\
\hline
Office-Home \cite{venkateswara2017deep} & 4 & 15500 & \makecell{Domains are: Art,\\Clipart, Product(object without\\background), Real-World.\\65 Classes are presented}\\
\hline
DomainNet \cite{peng2019moment} & 6 & \makecell{around\\600000} & \makecell{Images are from 345\\classes and domains are:\\clipart, real, sketch,\\infograph, painting,\\quickdraw.}\\
\hline
\end{tabular}}
\end{table}

\clearpage

\begin{landscape}
    \tiny
    \pagestyle{empty}
\setlength\LTleft{-60pt}            
\setlength\LTright{0pt}

\begin{longtable}{|c|c|c|c|c|c|}
\caption{\centering Review of Works - Domain Generalization \label{domgen}}\\
\hline
\textbf{Work}	& \textbf{Datasets}	& \textbf{Method Type} & \textbf{Method} & \textbf{Advantages} & \textbf{Disadvantages}\\
\hline
\cite{hu2021causal} & \makecell{Self-Defined} & \makecell{Novel\\(Causal)} & \makecell{Firstly, a graphical model is created to show relation\\between environment and agent, then after\\considering the local independencies, a hand-tailored\\loss is calculated and minimized.} & \makecell{Working on time-series,\\generalizable to other problems} & \makecell{Assumptions about Domain,\\lack of comparison to others}\\
\hline
\cite{wald2021calibration} & \makecell{Colored-Mnist,\\WILDS} & Information & \makecell{First, the paper proposes that a calibrated\\predictor on a few environment does not\\use any spurious features by providing\\mathematical intuition, then a few mechanism for\\model calibration are tested} & \makecell{Focusing on spurious correlation,\\can be applied on almost any model} & \makecell{Training environments require\\to be different to some extent} \\
\hline
\cite{cha2021swad} & \makecell{PACS, VLCS,\\OfficeHome,\\TerraIncognita,\\DomainBed} & \makecell{Novel\\(Optimization)} & \makecell{The idea is to seek flatter minima\\while minimizing the loss function of the model\\to improve the generalizability} & \makecell{Applicable to many tasks,\\simple intuition} & \makecell{Method weakness in finding flat minimas,\\weakness in utilizing to\\domain-specific information} \\
\hline
\cite{yang2021adversarial} & \makecell{PACS,VLCS,\\DomainNet,\\OfficeHome,\\Digit-DG} & \makecell{Novel\\(Knowledge\\Distillation)} & \makecell{An student network is trained on augmented data\\from other domain by a teacher seeing the main\\domain, meanwhile a generative model\\is trained for domain augmentation} & \makecell{Domain augmentation,\\ domain-robust representation} & \makecell{Leave-one-out evaluation is used\\which does not show model\\performance when training domain is fixed}  \\
\hline
\cite{zhou2020deep} & \makecell{Digits-DG,\\PACS,\\OfficeHome} & Generative & \makecell{A domain-transformation network is trained\\to create data from unseen domain and then\\synthetic data are used to train classifier} & \makecell{Domain-transfer network\\can also be used for data augmentation} & \makecell{Requires data from target domain}\\
\hline
\cite{domaingena1} & \makecell{ColoredMNIST,\\Camelyon17-WILDS,\\FMoW-WILDS,\\PACS} & Information & \makecell{First, the problem is transformed into\\an optimization problem where constraints\\are defined to consider information invariancy\\between domains, then a solution is proposed} & \makecell{Convergence guarantee,\\testing on breast cancer dataset} & \makecell{Complicated method which prevents\\others to improve upon} \\
\hline
\cite{huang2020self} & \makecell{PACS,VLCS,\\OfficeHome,\\ImageNet-Sketch} & Information & \makecell{Neurons with high-activations or high-gradient\\are shuted down while training} & \makecell{Ease of implementation, applicablity\\to many tasks} & \makecell{Neurons learn to correlate with most important\\features, and spurious correlation is a challenge}\\
\hline
\cite{ganin2016domain} & \makecell{Self-Defined,\\Amazon reviews} & Density & \makecell{Minimizing a loss to reach a representation of data\\where source and target distribution are indistinguishable} & \makecell{A pioneer, test on text\\image and synthetic data} & \makecell{Requires data from both domains}\\
\hline
\cite{sun2016deep} & \makecell{Office \cite{saenko2010adapting}} & Information & \makecell{Using linear transformation for aligning\\second-order statistics of distributions\\(for non-deep please refer to \cite{sun2016return})} & \makecell{A pioneer, ease of implementation} & \makecell{Linearity assumptions,\\requires data from target domain} \\
\hline
\cite{da2019style} & \makecell{Self-Defined} & Density & \makecell{Training a generative model for creating\\data from different path to generalize models\\for financial markets} & \makecell{Working On financial data,\\introduction of path as domain} & \makecell{Assumptions about domain,\\focusing On generative quality\\and not performance for generalization}\\
\hline
\cite{jin2021domain} & \makecell{Self-Defined} & Information & \makecell{Model aims to learn domain-invariant information\\by attention sharing while training a domain-discriminator\\and learning the domain specific information separately} & \makecell{working with time-series,\\considering scenarios such as cold-start} & \makecell{Only focusing on adaptation,\\only works on single-variate time-series} \\
\hline
\cite{guo2022evaluation} & \makecell{Medical\\Self-Defined} & Various & \makecell{Testing a few other model, not introducing One} & \makecell{Medical data, extensive evaluation and\\comparison of models} & \makecell{Assumption that data recorded in different\\years are from different domains}\\ 
\hline
\cite{li2017deeper} & \makecell{PACS,\\VLCS} & Information & \makecell{Undo-bias \cite{khosla2012undoing} idea is extended to NNs,\\then a low rank parameterization method\\is applied to reduce number of parameters} & \makecell{Introducing new benchmark,\\works better on high-shift datasets\\compared to prior works } & \makecell{Requires data from a few domains}\\
\hline
\cite{muandet2013domain} & \makecell{Synthetic,\\GvHD} & Information & \makecell{Firstly distributional variance (DV) is defined\\as a measure of divergence between domains,\\then an orthogonal transform is found from\\feature space to latent space which minimize DV} & \makecell{A pioneer, latent space transformation\\can be used to improve other algorithms} & \makecell{Requires data from a few domains}\\
\hline
\cite{wang2021learning} & \makecell{Digits,\\PACS,\\Corrupted\\CIFAR10} & \makecell{Generative\\Information} & \makecell{A generative model is proposed to generate\\data from different domains by minimizing the\\MI between generated data and samples\\(to diversify) and maximizing MI between samples of\\same class to keep semantics} & \makecell{Works on single\\domain generalization} & \makecell{Computationally expensive,\\comprations are not totally fair as\\other methods do not focus on\\single domain generalization} \\
\hline
\cite{cai2022face} & \makecell{CASIA-MFSD,\\Oulu-NPU,\\Replay-Attack,\\MSU-MFSD} & Generative & \makecell{A GAN model is improved for any-spoofing\\by forcing the discriminator to also\\distinguish between different domains} & \makecell{Simple intuition,\\improvement on mentioned task} & \makecell{Requires a few diverse domains\\to generalize enough for task\\at hand}\\
\hline
\cite{chen2022compound} & \makecell{PACS,\\Digits-DG,\\VLCS,\\OfficeHome} & \makecell{Information,\\Density,\\Novel} & \makecell{Firstly, a domain classifier is trained on some\\pseudo domain labels, which is used to normalize\\features, then semantic information is learned\\and extracted using prototypical relation modeling \cite{ding2021prototypical}} & \makecell{Does not require domain labels,\\A pioneer on using prototypical\\relation modeling} & \makecell{Requires data from a few domains,\\computationally expensive}\\
\hline
\cite{wan2022meta} & \makecell{Digits,\\CIFAR-10-C,\\PACS} & \makecell{Novel\\(Meta-Features)} & \makecell{A process is proposed to learn\\meta features (visual words) from convolutional\\features by minimizing a reconstruction loss and\\classify inputs based on best-match meta-features} & \makecell{Robust Representation of input data,\\Method can be also used\\to learn visual words for\\other OOD tasks} & \makecell{No constraint is introduced for\\disentanglement of meta features which\\might cause problem in domain shifts\\with huge change in distribution of classes}\\
\hline
\cite{zhang2022towards} & \makecell{Rotated MNIST,\\PACS,\\VLCS,\\Camelyon17-WILDS} & \makecell{Novel\\(Disentanglement)} & \makecell{Method tries to disentangle features showing\\semantic and variation information, former being\\extracted by an encoder trying to classify images\\ and latter by a decoder using different variation\\information to generate semantically same data} & \makecell{Does not require domain labels,\\Can be used to find variation\\between domains} & \makecell{Requires data from a few domains}\\
\hline
\cite{li2018learning} & \makecell{PACS,\\Synthetic} & \makecell{Information} & \makecell{In each training iteration, model\\is trained on a few of training domains\\aiming to improve on the rest of training\\ domains aiming to generalize better\\to unseen domains} & \makecell{Simple intuition,\\the idea can be applied\\ on almost any model} & \makecell{Requires data from a few domains}\\
\hline

\end{longtable}

\setlength\LTleft{-42pt}            
\setlength\LTright{0pt}

\begin{longtable}{|c|c|c|c|c|c|}
\caption{\centering Review of Works - OSR \label{osr}}\\
\hline
\textbf{Work}	& \textbf{Datasets}	& \textbf{Method Type} & \textbf{Method} & \textbf{Advantages} & \textbf{Disadvantages}\\
\hline
\cite{miller2021class} & \makecell{MNIST,\\SVHN,\\CIFAR10,\\CIFAR+10,\\TinyImageNet} & Density & \makecell{Introducing a loss which forces data\\of same classes to form tight clusters in latent space} & \makecell{Interpretability of latent space,\\model is appicalbe to any task} & \makecell{Requires training}\\
\hline
\cite{zhou2021learning} & \makecell{MNIST,\\SVHN,\\CIFAR10,\\CIFAR+10,\\TinyImageNet} & Novel & \makecell{Two models, one is a class placeholder, a neuron added to\\classification layer forcing the model to\\push the second-highest probability toward added neuron\\This is then used to distinguish in and out data\\ \& a data Placeholder, mimicing novel patterns with manifold\\mixup on latent embeddings} & \makecell{Simple intuition for class placeholders,\\small training overhead for\\class placeholder } & \makecell{Requires picking number of dummy\\classifier, data placeholder only working\\on embedding, requires training}\\
\hline
\cite{kong2021opengan} & \makecell{MNIST,\\SVHN,\\CIFAR10,\\TinyImageNet} & Generative & \makecell{A GAN is trained on feature level by an outlier exposed\\dataset, aming to learn a discriminator that can assign\\robust likelihood by considering the outlier data\\as fake when training GAN} & \makecell{GAN is trained on feature level\\thus the model is computationally\\efficient} & \makecell{Requires outlier exposed dataset\\for training}\\
\hline
\cite{vaze2021open} & \makecell{MNIST,\\SVHN,\\CIFAR10,\\CIFAR+10,\\CIFAR+50,\\TinyImageNet} & Various & \makecell{This Work focuses on showing that performance of\\any model in closed-set classes is\\correlated with its OSR performance.} & \makecell{Improving closed-set performance,\\introducing semantic shift benchmark,\\theorical justification with help\\of model calibration literature} & \makecell{Analysis of differences in shape of\\feature space between closed-set\\and open-set models is missing}\\
\hline
\cite{oza2019c2ae} & \makecell{MNIST,\\SVHN,\\CIFAR10,\\CIFAR+10,\\CIFAR+50,\\TinyImageNet} & \makecell{Reconstruction} & \makecell{An encoder-decoder network is trained, the\\encoder part is trained for classification \\ while the decoder used to calculate reconstruction loss\\and detect open set examples} & \makecell{Sample intuition, the idea is\\easily applicable to any\\encoder-decoder model} & \makecell{Requires threshold selection,\\classifier information is not used}\\
\hline
\cite{yoshihashi2019classification} & \makecell{MNIST,\\SVHN,\\CIFAR10,\\TinyImageNet,\\DBpedia} & \makecell{Reconstruction} & \makecell{A combination of openmax score is used with \\distance in latent space to detect OS samples;\\latent vectors are generated using a model that\\tries to reconstruct data with latent-vectors} & \makecell{Distance in latent space is used\\instead of reconstruction loss} & \makecell{Computation overhead for reconstruction\\part of model}\\
\hline
\cite{jain2014multi} & \makecell{Caltech 256+\\ImageNet,\\MNIST,\\LETTER} & Information & \makecell{A probabilistic model is introduced which\\fits a weibull distribution on decision score of \\SVM classifiers, and outputs probability for each class;\\a final thresholding detects OS samples} & \makecell{A pioneer, works on any model with\\score or probability output} & \makecell{Works on output of classification\\layer}\\
\hline
\cite{bevandic2022dense} & \makecell{Cityscapes,\\StreetHazard,\\Vistas,\\WildDash 1} & Novel & \makecell{A dense neural network is used for feature extraction;\\then features are used with labeled outlier data\\to train an OS detection network parallel\\to classification network} & \makecell{Working on semantic segmentation} & \makecell{Requires training,\\requires outlier data}\\
\hline
\cite{yang2019open} & MOCAP & Generative & \makecell{A GAN is trained to generate fake data\\which is used to train the classifier to\\distinguish known class from unknown} & \makecell{Tested on human activity recognition\\dataset} & \makecell{Requires training,\\working on output of classification\\layer} \\
\hline
\cite{scheirer2014probability} & \makecell{Caltech 256+\\ImageNet} & Information & \makecell{Similar to \cite{jain2014multi} with small changes\\in probability calculation} & \makecell{Extensive testing with different\\classifiers} & \makecell{Works on output of classification\\layer}\\
\hline
\cite{zhang2016sparse} & \makecell{YaleB,\\MNIST,\\UIUC,\\Caltech 256} & Information & \makecell{Extreme Value Theory is used to model tail\\distribution of data to test hypotisis of\\data belonging to train classes} & \makecell{Theoretical basis, superior results\\compared to other conventional ML models} & \makecell{Requires threshold selection,\\works on output of classification\\layer}\\
\hline
\cite{li2005open} & FERET & Information & \makecell{The classifier assigns a likelihood score, which\\is used to compute a peak-to-side ratio distribution\\(when samples are assigned to wrong classes);\\this distribution is used to find OS samples} & \makecell{A pioneer, works with many-classed\\problems (such as face recognition)} & \makecell{Approximations for likelihood score\\calculation, not tested against other\\means of scoring samples}\\
\hline
\cite{huang2022task} & \makecell{MiniImageNet,\\TieredImageNet} & Density & \makecell{This paper works on making GFSL \cite{gidaris2018dynamic}\\method Task adaptive by automatic threshold selection} & \makecell{Works on few-shot problems,\\threshold-free model} & \makecell{Requires OS data}\\
\hline
\cite{mundt2022unified} & \makecell{MNIST,\\FashionMNIST,\\AudioMNIST,\\KMNIST,\\SVHN,\\CIFAR} & \makecell{Generative,\\Information} & \makecell{Firstly, a $\beta$-VAE is trained on dataset, then a\\class conditioned EVT is applied on latent space\\to test whether data is inlier or OS} & \makecell{Workin on latent space instead of\\reconstruction error, EVT instead of\\threshold selection, extensive testing} & \makecell{Requires training, requires\\model selection for each dataset}\\
\hline
\cite{kim2021lane} & \makecell{Self-Defined\\and collected} & Density & \makecell{SVM algorithm is edited to counter for\\OS samples by EVT} & \makecell{Work on sensory data,\\work on lane change problem} & \makecell{Definitions of OS samples,\\method is not tested \\using benchmarks}\\
\hline
\cite{jang2022collective} & \makecell{Various\\Benchmark\\Datasets} & Information & \makecell{Firstly, it is shown that by replacing softmax\\with sigmoid, confidence scores are\\reduced for OS samples, then classifier\\is replaced with sigmoid-based one-vs-\\rest networks and using some rules,\\collective decision is drawn} & \makecell{Applicable to any classification network,\\only requires training of last layer} & \makecell{By increasing number of classes,\\computational demand also increases}\\
\hline
\cite{tran2022triple} & \makecell{MNIST,\\SVHN,\\CIFAR10,\\CIFAR+10,\\CIFAR+50} & Information & \makecell{A new activation is introduced which is a\\combination of 3 sigmoid functions which aims\\to tighten the boundary for inlier samples\\in classification layer to detect OS samples} & \makecell{Only changes last layer of network,\\can generate probability of OS} & \makecell{Requires hyper-parameter selection}\\
\hline
\cite{shao2022towards} & \makecell{Various\\touchless-\\biometric\\datasets} & \makecell{Density} & \makecell{In each iteration, a set of dataset is selected\\as support and another set as query, and meta\\feature extractors are updated distance of\\each query to respective meta sets} & \makecell{Works on feature layer,\\extracts robust features} & \makecell{Computationally expensive,\\not tested on benchmarks}\\
\hline
\cite{cai2022luna} & \makecell{ImageNet-LT,\\Places-LT,\\Marine\\Species-LT} & Density & \makecell{First, a loss function is added to\\decrease inter-class similarities,\\by forcing them to shape a compact cluster,\\then a novelty detection algorithm is proposed\\to find OS samples which also\\takes cluster size into account} & \makecell{Focusing on OSR in\\Long-Tailed recognition} & \makecell{Requires training}\\
\hline
\cite{yang2022one} & \makecell{SVHN,\\CIFAR10,\\CIFAR+10,\\CIFAR+50} & Information & \makecell{One neuron is added to the classification layer of\\network for OS samples; in training, two\\cross-entropy losses are minimized, one trying to match\\the output to coresponding class and second one\\matching the outputs after removing the neuron\\of correct class to OS neuron} & \makecell{Also tested for single domain\\generalization tasks,\\simple intuition} & \makecell{Requires training data\\to re-train the classifier} \\
\hline
\end{longtable}

\end{landscape}

While these datasets are the ones mostly used in research papers, they are mainly similar in the selection of domains and types of images. Some works have also presented datasets for other fields of study as well, such as satellite images \cite{christie2018functional} and breast cancer \cite{bandi2018detection}.

In DG problems, generalization to a new domain is the primary goal of papers; thus, any metric of performance in the new domain can be used for evaluation, from simple metrics such as accuracy to expected calibration error \cite{naeini2015obtaining}. By decreasing the number of domains used during training, the difficulty of the task at hand increases, the most common setting is one-vs-rest splitting (one domain for test, the rest for training). Meanwhile, the most challenging and also, at the same time, most interesting one is single-domain generalization. Also, if some training data from the test domain be available during training, the problem changes to domain adaptation \cite{wang2018deep}, which is not within the scope of our study. Lastly, table \ref{domgen} reviews work done on domain generalization.

\subsection{Open Set Recognition (OSR)}

Open set recognition is the task of detecting data from new/unseen classes. The possibility of new events is inevitable in open-world settings, as it is impractical to account for any probable event during the creation and training of the model. This problem can be viewed as shifts in $P(Event)$.

Creating datasets for OSR model evaluation can be easily archived by removing one or more classes from the dataset during training and then considering it back during the test. To compare datasets, Scheirer et al. \cite{scheirer2012toward} suggested openness which aims to measure how open the underlying problem is. The following equation is used to calculate this metric:
\begin{equation}
    Openness = 1 - \sqrt{\frac{2*N_{Train}}{N_{Test}+N_{Target}}}
\end{equation}
Where $N_{Train}$ is the number of classes used during training, $N_{Test}$ is the number of classes that will be observed during testing, and finally, the number of classes which needs to be correctly recognized in testing as $N_{Target}$. In addition, to randomly remove some classes during training, to increase openness without further reducing the number of training classes, in CIFAR+$N$ datasets, some non-overlapping classes are added to CIFAR10 from CIFAR100.

In this problem setting, the main goal is to detect open set (OS) samples while making the fewest possible mistakes on in-lier data points. Therefore, two sets of metrics need to be defined, the first one evaluating performance on closed-set data (usually accuracy) and the second one for measuring OSR performance. However, the ratio of OS samples in test time is unknown; thus, every algorithm requires a sensitivity parameter to be set \cite{neal2018open}, which makes the area under the ROC curve (AUROC) fitted for the job. Finally, table \ref{osr} reviews work done on OSR.

\subsection{Outlier Detection and Adversarial Examples}

Moving to changes caused by shifts in $P(Noise)$, outlier detection and adversarial examples are the next category we check. 

In an outlier detection setting, compared to other problems, usually, what causes the OOD sample is not a shift in $P(Noise)$ but the occurrence of an unlikely value. This has caused a problem in dataset creation in papers in this field, as many of them merely apply an OSR setting to their model and evaluate the model on their ability to recognize images from unseen classes. Nevertheless, server machine dataset (SMD) \cite{su2019robust} is commonly used among researchers in this field, and many have worked on time-series and online outlier detection, which are the current trends.

Next, adversarial examples are one of the best-known types of distribution shifts that affect the performance of deep neural networks dramatically. These data points are usually generated by adding a small perturbation to data; usually, these types of examples are not separable from main data by the human eye. These types are named adversarial examples, as they can be used for adversary purposes \cite{goodfellow2014explaining}.

Given the nature of challenges and limitations in attacking any system and the final goal of the attacker, there are plenty of possible adversary attacks. The underlying system might be a black box with only a limited number of possible accesses to it, or the data might go under preprocessing before being fed to the ML system. Many surveys \cite{chakraborty2018adversarial,xu2020adversarial} have reviewed the adversarial examples and attacks on DL and ML systems on different types of data, and the interested reader can check them for further details on the topic. There are also many famous attacks such as L-Broyden–Fletcher–Goldfarb–Shanno (L-BFGS) \cite{szegedy2013intriguing}, fast gradient sign method (FGSM) \cite{goodfellow2014explaining} and basic iterative method (BIM) \cite{kurakin2018adversarial}, and it is common to evaluate models against a few of them.

As the models' main aim here is to recognize outliers, evaluation metrics are similar to OSR. Also, for these two problem settings, there are two great python libraries: pyod \cite{zhao2019pyod} for outlier detection, and ART \cite{art2018} for adversarial attacks. These libraries have implemented many of the algorithms in a plug-and-play fashion, which can be genuinely valuable to researchers. Table \ref{novel} reviews work done on outlier and adversarial attacks detection.


\subsection{General OOD}

As we categorize each problem and shift by the reason behind it, the inevitable question is, \textit{what if more than one cause is behind the change in the distribution of data?}

For example, consider a model trained on sketches of cars to be evaluated on real photos of all types of vehicles. General out of distribution detection, or OOD detection, aims to detect OOD samples in any case rather than merely focusing on a specific type of change. In this setting, generalization is also meaningless, given that the OOD sample might be from new events or even an outlier. Notably, adversarial attacks are usually not considered in OOD literature, given their difference in nature (intentional vs. unintentional changes).

Also, it is noteworthy to mention that general OOD problems are not always harder than other categories. They are also easier to define in new settings, given that they do not require the assertion of the cause of changes to be merely one of the environment, event, or noise.

In this setting, a usual approach is to consider one dataset as in-liers and another one as outliers. Mostly, the challenge in this setting is to find methods to achieve OOD detection without training a model and using an already learned model; thus, it is common to consider image net or imagenet1k as an inlier dataset. This selection is due to the fact that for most DL models, pre-trained networks on imagenet are available. Also, having outlier samples (mainly to see how the model behaves) or not can affect the difficulty of the problem at hand.

To evaluate these models, similar to the OSR setting, AUROC is commonly used in research papers. The false-positive rate at 95\% true positive rate is also widely seen in research papers. A simple intuition behind this metric is setting the sensitivity on the model in a way to pass 95\% of positive data and then measuring how many outliers have passed the filter. Finally, researchers also use the area under the precision-recall curve (AUPR) as a metric. Lastly, table \ref{ood} reviews work done on General OOD detection.

\clearpage

\begin{landscape}
    
    \scriptsize
    \pagestyle{empty}
\setlength\LTleft{-100pt}            
\setlength\LTright{0pt}

\begin{longtable}{|c|c|c|c|c|c|}
\caption{\centering Review of Works - Outlier and Adversarial Attacks Detection \label{novel}}\\
\hline
\textbf{Work}	& \textbf{Datasets}	& \textbf{Method Type} & \textbf{Method} & \textbf{Advantages} & \textbf{Disadvantages}\\
\hline
\cite{cai2021learned} & \makecell{Synthetic,\\VIRAT Video,\\Ultrasound} & Reconstruction & \makecell{A method is proposed to learn RPCA with\\neural networks, which is used\\to detect outliers} & \makecell{Explainability of learnt parameters,\\speed compared to other works} & \makecell{Requires stopping criteria,\\requires hyperparameter selection} \\
\hline
\cite{rebjock2021online} & \makecell{Synthetic,\\SMD} & Information & \makecell{This paper focuses on solving $\alpha$-death\\problem by introducing a memory decay term to\\existing methods} & \makecell{Online outlier detection,\\works on few-outlier environment} & \makecell{Requires hyper-parameter selection} \\
\hline
\cite{su2019robust} & \makecell{SMD,\\SMAP,\\MSL} & Reconstruction & \makecell{An RNN is combined with a VAE to learn robust\\representation of data, then reconstruction error\\is used to detect outliers} & \makecell{Variational representation learning,\\entry-level outlier detection on\\multivariate time-series} & \makecell{Complexity of model} \\
\hline
\cite{tack2020csi} & \makecell{CIFAR10,\\CIFAR100,\\Imagenet30} & \makecell{Novel\\(Self-Supervision,\\Representation)} & \makecell{OOD-ness of a few augmentation techniques are tested\\with discriminative ability of model after shift, then\\hardest are picked as main shift is encoded\\in SimCLR\cite{chen2020simple} loss to learn robust representation} & \makecell{Robust representation to discriminate\\in and out distribution, improvement on harder\\OOD samples} & \makecell{Testing shifts are created and\\picked by experts knowledge}\\
\hline
\cite{wang2020further} & \makecell{CIFAR10,\\CelebA,\\TinyImageNet,\\SVHN} & Density & \makecell{White noise test is used as a weak relaxation of\\IID-ness to detect outliers with generative models} & \makecell{Unsupervised, possibility of incorporating prior\\knowledge about outlier distribution,\\Analysis on hard scenarios,\\where textual differences is minimized} & \makecell{The underlying assumption is too weak, given\\that White Noise is weaker than\\martingale difference which is itself\\weaker than IID}\\
\hline
\cite{yang2021mean} & \makecell{Countries,\\unbalance,\\and XOR from\\sipu datasets} & Density & \makecell{Mean-shift algorithm with an outlier\\filtering step is used to cluster data and find\\outliers} & \makecell{Works on many-outlier environments,\\fast} & \makecell{Not tested on high-dimensional data\\like images}\\
\hline
\cite{cheng2021unsupervised} & \makecell{MNIST,\\Fashion-MNIST,\\SVHN,\\CIFAR-10,\\CIFAR-100} & \makecell{Reconstruction} & \makecell{Images are transformed with data augmentation\\techniques before feeding to an autoencoder\\which aims to learn reconstruction while\\being invariant to transformation} & \makecell{More robust considered to simple\\reconstruction methods, trained\\ with adaptive self-paced learning} & \makecell{Requires threshold selection,\\only works in environment where\\transformations are defined}\\
\hline
\cite{li2020copod} & \makecell{30 different\\from ODDS\\and DAMI \cite{campos2016evaluation}} & Density & \makecell{First, empirical copula is used to fit\\a distribution to dataset, then the tail\\probability is calculate to reject or accept samples} & \makecell{Parameter-free, interpretability,\\works on high-dimensional data} & \makecell{Empirical copula might require\\a lot of data to fit a\\distribution properly}\\
\hline
\cite{cheng2021learning} & \makecell{Stanford Dogs,\\FounderType-200,\\CUB-200-2010} & Density & \makecell{A loss function is proposed to\\force feature space to have a\\class-conditional gaussianity shape} & \makecell{Working on feature space,\\proposed loss can be used\\for other tasks as well} & \makecell{Computational overhead, only\\appicalbe to neural networks}\\
\hline
\cite{shah2021three} & \makecell{20 Newsgroups,\\50 Class Reviews} & Density & \makecell{For each class of data, the classifier\\devides the space into 3 parts, namely,\\inside, outside and partial which\\forces the model to learn more compact\\sub-spaces} & \makecell{Simple intuition,\\can be changed into a loss\\function for neural networks} & \makecell{Working on input space,\\test settings are similar to\\OSR which makes comparison with other\\novelity detection methods invalid}\\
\hline
\hline
\cite{wang2021revisiting} & \makecell{MNIST,\\CIFAR-10,\\CIFAR-100} & Information & \makecell{Hilbert Schmidt Independence Criterion (HSIC) is used to\\estimate mutual information, and information bottleneck\\is introduced as a regularization term to make model\\robust against adversarial examples} & \makecell{Works without adversarial examples\\during training, low drop in\\discriminative performance} & \makecell{Computational overhead, using HSIC\\to estimate mutual information is not\\a new idea}\\
\hline
\cite{lee2018simple} & \makecell{SVHNN,\\CIFAR-10,\\CIFAR-100} & Density & \makecell{A Gaussian classifier is used to\\compute confidence scores using\\Mahalanobis distance in each\\layer of network} & \makecell{Better AUC compared to softmax,\\introduction of an algorithm\\for calibration} & \makecell{Requires training data,\\requires to re-feed all data\\to network}\\
\hline
\cite{zheng2018robust} & \makecell{MNIST,\\F-MNIST,\\CIFAR-10} & Generative & \makecell{A generative model is used to model the \\distribution of hidden states (hidden neurons) of model\\which is later used to detect adverarial examples} & \makecell{Works on feature space, tested\\againsed FGSM and DeepFool} & \makecell{Requires training data,\\requires to re-feed all data\\to network} \\
\hline
\cite{massoli2021detection} & \makecell{VGGFace2,\\MNIST,\\CIFAR-10} & Density & \makecell{Firstly, using the inlier data, class\\centroids for each layer of network is calculated,\\then detectors are trained to find\\adversary examples using distances to\\these centroids} & \makecell{Does not require training for\\base model, generalizability between\\differet problem settings} & \makecell{Requires adversarial training}\\
\hline
\cite{wang2021multi} & \makecell{Market1501,\\DukeMTMC-ReID} & Density & \makecell{Firstly, a few expert models are selected,\\then it is shown empirically than by analysing first K\\ outputs of models, a difference is seen in\\query-support relations, support-support relations\\and cross-expert relations} & \makecell{Focusing on re-identification,\\multi-modal (compared to \\other single-model works)} & \makecell{Working on output of classification\\layer of models,\\detector requires adversarial training} \\
\hline
\end{longtable}

\setlength\LTleft{-70pt}            
\setlength\LTright{0pt}

\begin{longtable}{|c|c|c|c|c|c|}
\caption{\centering Review of Works - OOD detection \label{ood}}\\
\hline
\textbf{Work}	& \textbf{Datasets}	& \textbf{Method Type} & \textbf{Method} & \textbf{Advantages} & \textbf{Disadvantages}\\
\hline
\cite{serra2019input} & \makecell{A few\\vs CIFAR10} & Density &  \makecell{Combining an upper-bound on Kolmogorov complexity\\with likelihood From generative models} & \makecell{Ease of implementation,\\Works with any generative model that\\assigns likelihood} & \makecell{Using compression for guessing\\the complexity of data}\\
\hline
\cite{morningstar2021density} & \makecell{Various Datasets\\Such as MNIST\\vs Each Other} & \makecell{Density/\\Information} & \makecell{Instead of comparing model probabilities,\\statistics from model probabilities are\\calculated to fit a model on density of them\\(by kernel density estimator or SVM), which is\\then used to detect OOD samples} & \makecell{Does not require labeled\\data or OOD samples, applicable\\to any trained model} & \makecell{Requires threshold selection,\\complicated mathematical theory\\prevents other studies to build\\upon this one} \\
\hline
\cite{liang2017enhancing} & \makecell{CIFAR-10 \&\\CIFAR-100 vs\\TinyImageNet,\\LSUN} & Information & \makecell{Small perturbations are added to increase the softmax\\scores, then it is observed that this\\affects in-liers more than OOD smaples\\which is used to detect them} & \makecell{Used widely as baseline in\\other papers, works with any\\neural network} & \makecell{Requires threshold selection,\\works on output of classification\\layer, computational overhead of\\perturbation calculation}\\
\hline
\cite{sun2021react} & \makecell{ImageNet-1k vs\\Places365,\\Textures,\\iNaturalist,\\SUN} & Information & \makecell{The activation's of penultimate layer outputs are\\truncated to limit the effect of noise,\\then an scoring function is applied\\to find OOD samples} & \makecell{Applicable to any pre-trained networks,\\can work with any scoring function\\such as ODIN\cite{liang2017enhancing}} & \makecell{Requires threshold selection,\\requires output of penultimate layer for\\training samples}\\
\hline
\cite{gawlikowski2022advanced} & \makecell{AID,\\UCM,\\LCZ42} & Information & \makecell{A Dirichlet Prior Network (DPN) is trained\\with two losses, one for in-data\\which aims to increase performance and another\\for OOD data to force an uniform output} & \makecell{Works in many-class problems,\\Tested on remote sensing data,\\test designs} & \makecell{Requires Training and OOD data}\\
\hline
\cite{ren2019likelihood} & \makecell{Fashion-MNIST\\ vs MNIST,\\CIFAR-10\\vs SVHN,\\Self-Defined} & \makecell{Generative,\\Information} & \makecell{The idea is to cancel effect of\\background in likelihood calculation} & \makecell{Working on gnomic data,\\Visualization and interpretation\\of results} & \makecell{LSTM model is tested for\\DNA data generation, which\\is not intrinsically generative,\\Requires Training} \\
\hline
\cite{liu2020energy} & \makecell{SVHN,\\CIFAR-10,\\CIFAR-100\\vs SVHN,\\Textures,\\Places365,\\LSUN,\\iSUN} & Information & \makecell{An energy score is calculated using\\probabilities provided by softmax output\\to find OOD samples} & \makecell{Simple intuition, no need to\\train networks} & \makecell{Works on output of classification\\layer, Requires threshold selection}\\
\hline
\cite{luan2021out} & \makecell{MNIST,\\GTSRB} & Density & \makecell{Isolation forest (IF) outlier detection is applied to mid-\\level features of network to find OOD samples} & \makecell{Network does not require\\further training} & \makecell{Requires in-data to train IF, IF\\may not perform well on networks\\with high-dimensional layers, not\\tested against other state-of-the-art OOD\\detection methods}\\
\hline
\cite{nitsch2021out} & \makecell{KITTI vs\\NuScenes} & Generative & \makecell{A GAN is trained to create fake data\\similar to in-data, then a classifier is trained\\to generate uniform output for out-data\\and class label for in-data} & \makecell{Dataset used for evaluation\\of model is new} & \makecell{Data sets might also include\\domain shift, GAN might create in-\\data}\\
\hline
\cite{papadopoulos2021outlier} & \makecell{CIFAR-10,\\Places365} & Information & \makecell{Classification layer is trained to minimize\\total variation distance between output of\\outlier data and uniform distribution} & \makecell{Using new distance metric\\compared to other works} & \makecell{Requires outlier data,\\requires training\\(at least for classifier)}\\
\hline
\cite{roy2022does} & \makecell{Dermatology\\Images\\Self-Defined} & Information & \makecell{Method consists of two stages, firstly,\\using four different pre-trained Resnet101s,\\representation of images in feature space is calculated,\\then a fine-to-coarse classifier is trained to classify\\inlier and outlier data and their classes} & \makecell{Working on medical data,\\ focusing on small shifts caused by\\changes in condition} & \makecell{Requires some outlier for training,\\makes an assumption about\\having label for outlier data}\\
\hline
\cite{zaeemzadeh2021out} & \makecell{CIFAR-10 \& \\CIFAR-100 vs\\ TinyImagenet,\\LSUN - \\UCF101,\\HMDB51} & Density & \makecell{Features from layer before last FC are forced to\\shape as a union of 1d sub-spaces, then\\at test, spectral discrepancy between sample\\and vectors for predicted class\\are used to detect OOD} & \makecell{Working on Feature space\\} & \makecell{Requires Training data,\\Latent Space Size can effect performance}\\
\hline
\cite{morteza2022provable} & \makecell{CIFAR-10 \&\\CIFAR-100 vs\\Textures,\\SVHN,\\Places365,\\LSUN,\\iSUN} & Information & \makecell{A Gaussian distribution is fitted on outputs of\\feature layer of network which is used to\\find in-distribution probabilities} & \makecell{Network does not require\\furthur training} & \makecell{Requires in-data to fit Gaussian,\\Requires threshold selection} \\
\hline
\cite{wang2022partial} & \makecell{CIFAR10-LT,\\CIFAR100-LT,\\Imagenet-LT vs\\a few benchmarks} & Density &\makecell{A Contrastive learning loss is edited to merely\\pull distance between out-data and tail in-data\\without considering head in-data} & \makecell{Focusing on OOD in\\Long-Tailed recognition} & \makecell{Requires out-data during training}\\
\hline
\cite{lust2022efficient} & \makecell{Various\\behchmarks\\against\\each other} & Information & \makecell{First, label of input is found by feeding it\\to network. Then a smoothed version of image is\\fed to network and an uncertainty score is calculated\\based on its output and label of image\\which is used to detect OOD} & \makecell{Tested on various settings\\such as adversarial sample\\detection and novel corruption,\\does not require training} & \makecell{Works on output of\\classification layer}\\
\hline
\end{longtable}

\end{landscape}

\section{Challenges and Future Directions}

In this section of the paper, we discuss our review's findings and the challenges that still need to be addressed in future studies. While the last sections of the paper were mainly devoted to reviewing each work separately, here, we aim to discuss their findings as a whole to pave the path for new researchers in the field.

\begin{figure}[h]	
\centering
\includegraphics[width=\textwidth]{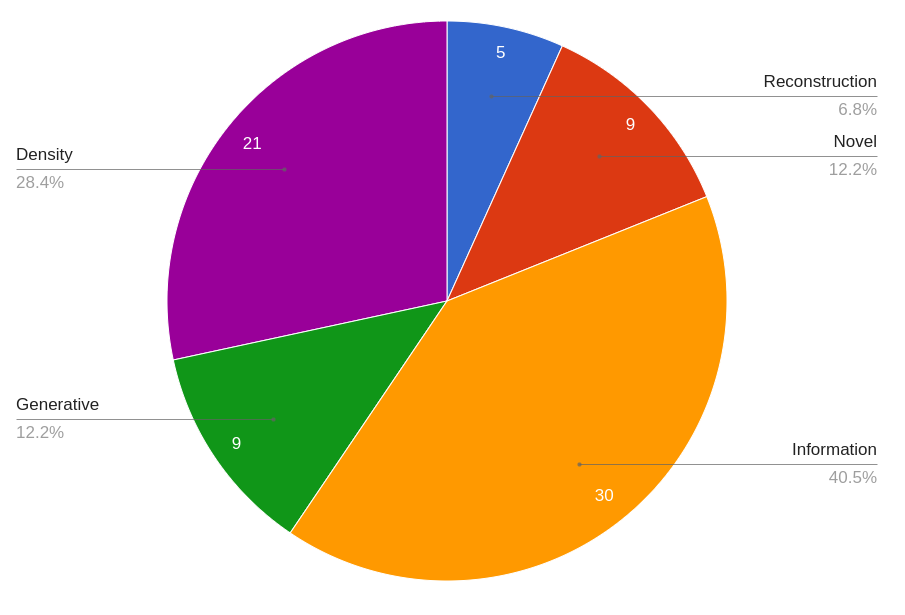}
\caption{Various types of methods and frequency of their appearance\label{freq}}
\end{figure}

Firstly, after looking at figure \ref{freq}, it is observable that the most common type of method for OOD detection in papers is information theory backed ones. This is also a logical observation, given that these methods do not require training a new model compared to reconstruction or generative ones. As a result, this trend is expected to continue in the future. Also, Novel models have started appearing in the past couple of years, and their potential yet needs to be explored.

As for implementation, code availability, and testing, some libraries exist for each of these problem settings, and Github repositories of many of them are also available publicly. Consequently, code availability and result reproduction are not mainly a problem in this field of study; however, more recent papers focus on large datasets and models, which makes testing their models nearly impossible for individual researchers. Working on more compact models and benchmarks can help researchers dramatically.

A few challenges still need to be addressed in evaluation and test settings. It is not easy to compare papers based on their findings without a unified protocol. A complete protocol must contain specific information about the dataset, its splits, metrics, and experiments. This can be viewed to some extent in papers, especially in the ones published in top-tier conferences and on domain generalization, and they mostly use the same protocol as their prior; nevertheless, designing a comprehensive experiment setting can be the focus of a future study.

Next, mathematical analysis of the difficulty of tasks needs to be addressed. In OSR, the openness of a dataset is introduced as a metric to measure the difficulty of the task at hand; the missing of a similar measure can be felt in other problem settings. Moreover, for some problems such as domain generalization and OSR, other qualitative measures of difficulty can be introduced as well, as generalizing from real photos to sketches is usually considered a more challenging task than from quick-draw to sketches.

Given the nature of OOD detection, it is expected for models which learn with fewer data to perform better. This can also be seen in model designing, as most of the methods currently do not limit themselves to a partial observation of training data (or no data at all). Moreover, some methods still use outlier data to train their model, which can be problematic when facing different behavior in OOD samples.

The type of data used is also another challenge faced in OOD literature. As it is shown in the tables, image data is the most commonly used type of data, and others have been left behind. The introduction of datasets on time series, videos, and texts is a direction for future work, which can unveil other challenges that have not been seen yet.

Long-Tailed recognition \cite{liu2019large} has also emerged as another field of study that addresses recognition problems in classes with limited data points. Investigation of the performance of models in this setting and in few-shot learning problems is another direction for future works; notably, a few works have recently initiated this path, such as \cite{cai2022luna}.

Explainability \cite{angelov2020towards} and interpretability \cite{zhang2018visual} of models is another direction toward trustworthy AI, which focuses on making decisions of model rational. Investigation of how these models perform in OOD settings can be a direction for future studies. Also, these models can be used to explain how OOD detection methods work to improve them further.

Finally, as Occam's razor rule mentions, the best model is the simplest one. Nevertheless, simple intuition is not the strongest suit of OOD detection methods, and another direction for researchers can be working on the simplicity of models instead of improving the results of prior ones.

\section{Discussion and Conclusion}
In recent years, AI has dramatically changed the world and affected every human's daily life. From voice assistants \cite{hsieh2021hey}, and autonomous cars to scales of smart education and smart cities \cite{lee2020artificial}, traces of artificial intelligence can be seen almost everywhere. Even now, it can be predicted that applications of AI are still expecting exponential growth in their capabilities \cite{calum2018artificial}; hence, another upsurge in their usage is anticipated. These methods are also expected to perform well in their tasks, as they are used in many critical sectors such as healthcare and military \cite{masuhr2019ai}. Moreover, some errors in the output of AI-based methods can also suffer from delayed error detection, such as mobile healthcare \cite{yang2022review} where a patient might postpone his doctor appointments by getting wrong healthy results from his device.

After years of research, there is still a visible gap between most of the research papers and real-world scenarios. Rigorously saying, in the field of ML, many of the papers suffer from proper test designing. The problem arises from the nature of the metrics used in these works, which are all based on the model's performance in a closed-world, pre-defined, identically and independently distributed dataset. In reality, none of these three conditions can be guaranteed to hold in many real-world problems. Nevertheless, after being violated in new environments, current models perform sensibly poorly in their tasks. These problems, while essential to solve, can also be seen as a new field of research that is built upon the prior ones, a path aiming to make previously created models applicable in a realistic setting. 

Before moving to solve any problem, it is necessary to define expectations from the outcome of the proposed solutions. In OOD literature, the most straightforward definition of expectation from models is to perform well in any situation by assigning correct labels to known classes of data and detecting the unknown. However, this simplest definition is also the most challenging level to achieve. In the simplest form, we can expect models to detect any data which is from a different distribution than training data, and in a more complicated form, we might expect the model to generalize to new data; where some assumptions should be held, such as having enough information or diversity in training data, so the task becomes reasonable. Even after this, preventing models from relying on spurious correlations is challenging. Also, everyone should keep in mind that making mistakes is not the same as performing poorly, and any performance should be measured and evaluated by allowing a fraction of outputs to be wrong; this is rational given that humans also make mistakes all the time, and we still have a long way before competing against human benchmarks.

Until now, we have discussed the importance of the problem at hand and what we expect from the solutions. The next logical topic of discussion is neighbor fields of study; and what makes them different. From one aspect, OOD literature contributes toward making trustworthy AI models, which makes it a neighbor with calibration \cite{krishnan2020improving} and reliability \cite{balagurunathan2021requirements} research from a quantitative view and with model interpretation \cite{zhang2018visual} and explainability \cite{angelov2020towards} from a qualitative one. Moreover, from a mathematical view, OOD literature shares a considerable amount of ideas with works done on information and probability theory literature, such as distribution matching \cite{schulte2015constant}. Lastly, OOD takes advantage of many recent ideas and trends in deep learning literature, such as disentangle representation learning \cite{liu2022learning} and causal learning \cite{scholkopf2021toward}.

In this paper, we have reviewed more than 70 papers from different segments of OOD literature. In order to form this review, we have first shown how the literature is segmented into different problem settings and what makes each of them unique. By using an SCM, we have tried to clarify the differences and show the reasoning behind every type of change in data distribution. This helps researchers quickly find the best match for their application, given the behavior of their underlying environments. Moreover, we have discussed categories and types of methods applied in the literature. By reading sections III and IV of our paper, interested readers can get a short introduction to OOD literature, its segments, and proposed solutions.

Next, we reviewed papers in detail in section 5. For each problem setting, a short discussion is presented, which gives enough information and intuition about the problem, benchmarks, and toolboxes that can also work as a head-start for researchers seeking a starting point for this field. Afterward, papers are reviewed in detail in a table for each segment; tables contain information about the type of method, a short summary of it, its advantages and disadvantages, and finally, used datasets. While some might argue that the dataset column is unnecessary in a review table, this can help researchers quickly find works to compare their methods.

Looking at the tables, the first visible observations are about the types of methods used. Definitely, information theory based methods are the most common in all different categories of shifts. However, while comparing different problem settings, the second most used type of solution is usually unlike. As domain generalization is the only problem setting that pushes toward generalization (rather than detection), studies mainly use novel methods. The reasoning behind that can be seen in the nature of other types of solutions, which are intrinsically created for detecting and recognizing OOD data points. However, density and reconstruction-based methods are more popular in the rest of the problem categories. Also, generative methods are more often seen in recent papers; arguably, they have not yet been appropriately evaluated for the task at hand, but some ideas, such as domain augmentation, are worthy of mentioning and probably will be more analyzed in the future. Overall, for the choice of method, the best and most obvious one is the information theory backed ones, and after that, density and reconstruction based ones for detection and novel methods for generalization.

Next, by looking at the advantage and disadvantage columns of the tables and simultaneously at the publication date of papers, we can observe that most of the papers can be considered as works in parallel with each other. This might help the literature by diversifying the solutions but causes a delay in solving the main problem. One solution to remedy this issue is introducing new challenges and benchmarks to help researchers compete against each other and also make them informed about the works of others faster. Moreover, in each problem setting, one common disadvantage of papers is focusing on easier problems and benchmarks, which makes their models not comparable with state-of-the-art methods. This is also resolved to some extent with the introduction of new benchmarks. Finally, a detailed analysis of the advantages and disadvantages of reviewed works is presented in section 6 of the manuscript, which aims to introduce directions for future works.

Lastly, as mentioned, we presented challenges left for future works in section 6; by analyzing the works reviewed, we have tried to find places where the research is mostly missing and needed and suggest them as directions for future research. While we have tried to make this section as comprehensive as possible, the possibilities for future works are limitless; thus, the readers seeking ideas are encouraged to look at the disadvantage column of tables.

To sum it up, we hope that our survey works as a starting point for new researchers in the field and provides a unifying look into the problem at hand for anyone working on OOD literature. Even though handling distribution shifts in ML literature is not new, works on more advanced DL techniques have started merely recently, and the field can be considered somewhat young. Hopefully, the next few years will hold many advancements for OOD detection, and many challenges will be addressed, making the models that we use in our everyday life more trustworthy.

\bibliography{main}

\end{document}